\documentclass[a4paper,fleqn]{cas-dc}

\usepackage[numbers]{natbib}
\setcitestyle{square,numbers,sort&compress}
\usepackage{calligra}
\usepackage{algorithm}
\usepackage{algorithmicx}
\usepackage{algpseudocode} 
\usepackage{amsmath}
\usepackage{float}
\usepackage{caption}

\def\tsc#1{\csdef{#1}{\textsc{\lowercase{#1}}\xspace}}
\tsc{WGM}
\tsc{QE}
\tsc{EP}
\tsc{PMS}
\tsc{BEC}
\tsc{DE}

\begin{document}
\let\WriteBookmarks\relax
\def\floatpagepagefraction{1}
\def\textpagefraction{.001}

\shorttitle{}

\shortauthors{}

\title [mode = title]{CtrlNeRF: The Generative Neural Radiation Fields for the Controllable Synthesis of High-fidelity 3D-Aware Images.}                      
%
\author[1]{Liu, Jian}[type=editor,
                        style=chinese,
                        auid=000,bioid=1,
                        orcid=0000-0003-2981-2128]

\cormark[1]

\ead{liujian10@zzu.edu.cn}

\credit{Conceptualization of this study, Methodology, Programming}

\affiliation[1]{organization={School of Computer and Artificial Intelligence, Zhengzhou University},
    addressline={No.100 Science Avenue}, 
    city={Zhengzhou},
    state={Henan},
    postcode={450001},
    country={China}}
   
\author[2]{Yu, Zhen}[%
    style=chinese,
    auid=000,bioid=1
   ]

\credit{Writing - Original draft revision}

\affiliation[2]{organization={Department of Electrical and Computer Engineering, California State Polytechnic University},
    city={Pomona},
    postcode={CA 91768},
    country={USA}
    }

\begin{abstract}
The neural radiance field (NERF) advocates learning the continuous representation of 3D geometry through a multilayer perceptron (MLP). By integrating this into a generative model, the generative neural radiance field (GRAF) is capable of producing images from random noise $z$ without 3D supervision. In practice, the shape and appearance are modeled by $z_{s}$ and $z_{a}$, respectively, to manipulate them separately during inference. However, it is challenging to represent multiple scenes using a solitary MLP and precisely control the generation of 3D geometry in terms of shape and appearance. In this paper, we introduce a controllable generative model ($i.e.$ \textbf{CtrlNeRF}) that uses a single MLP network to represent multiple scenes with shared weights. Consequently, we manipulated the shape and appearance codes to realize the controllable generation of high-fidelity images with 3D consistency. Moreover, the model enables the synthesis of novel views that do not exist in the training sets via camera pose alteration and feature interpolation. Extensive experiments were conducted to demonstrate its superiority in 3D-aware image generation compared to its counterparts.
\end{abstract}


\begin{keywords}
Implicit Representation\sep Novel View Synthesis\sep Generative Adversarial Network (GAN)\sep Neural Radiation Field (NERF)\sep Controllable Image Generation \sep 3D-Aware Images
\end{keywords}
\maketitle
\section{Introduction}
\begin{figure}[htbp]
\centering
\includegraphics[width=6cm]{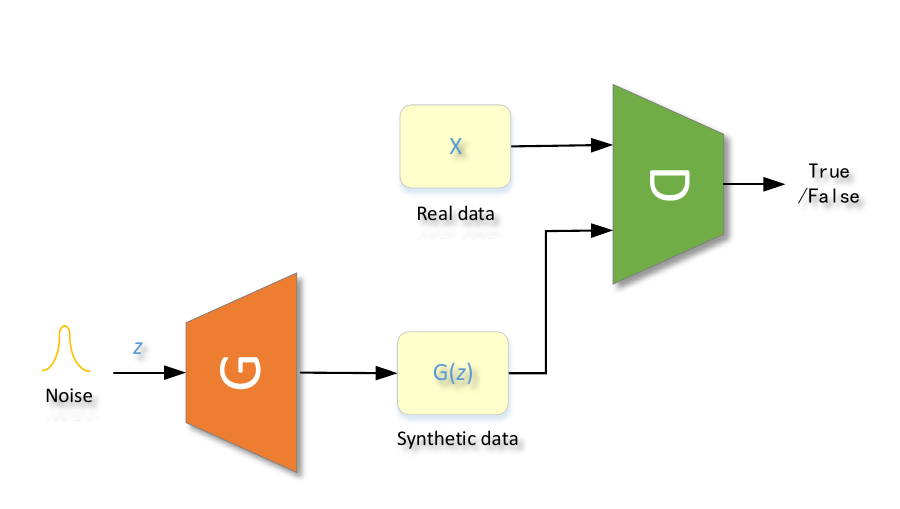}
\caption {The architecture of generative adversarial networks (GANs). G refers to a generator, and D refers to a binary discriminator.}
\label{fig:1}
\end{figure}

\begin{figure*}[htbp]
	\centering
		\includegraphics[width=13cm]{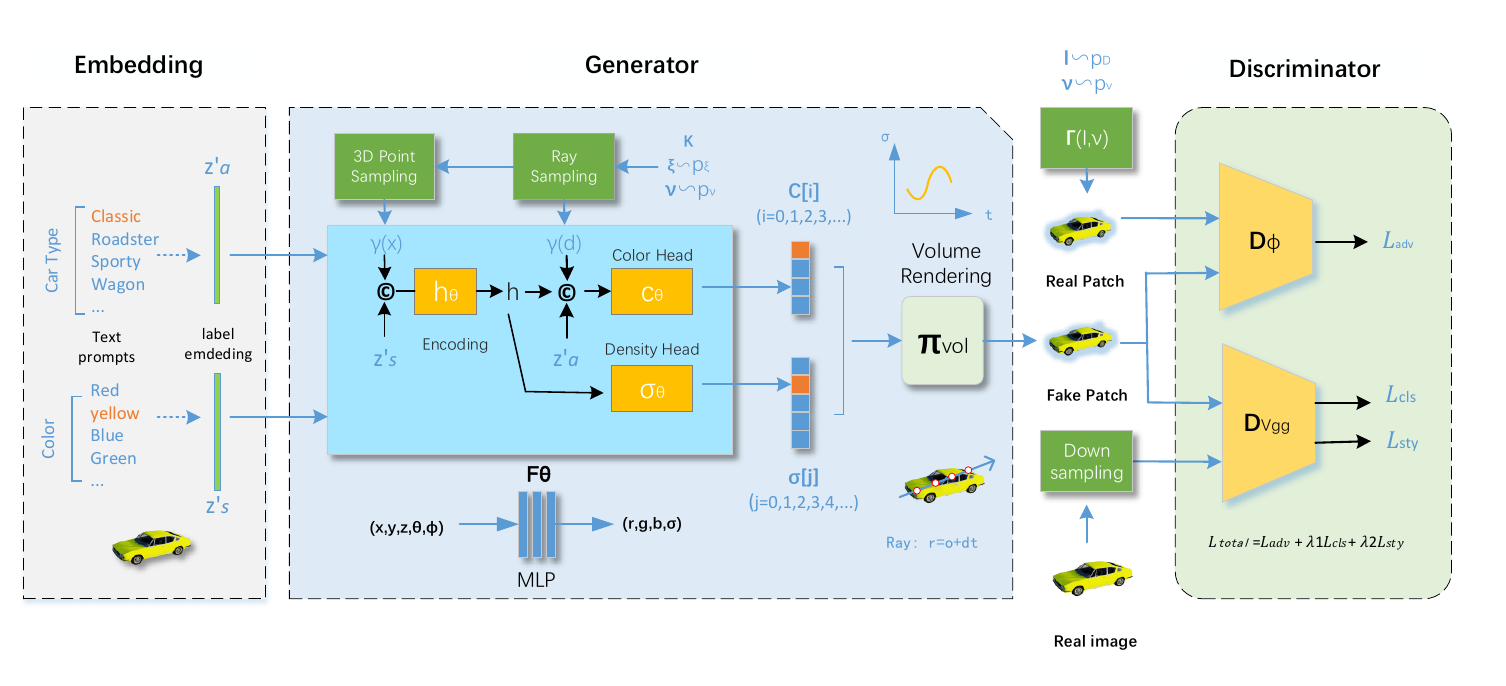}
	\caption{The framework of the generative neural radiation field (CtrlNeRF), which includes three main components: embedding, generator, and discriminator.}
	\label{fig:2}
\end{figure*}

\indent In 2014, Goodfellow et al. proposed a generative adversarial network (GAN) \cite{2014Generative}, which is a deep generative model inspired by game theory. Subsequently, various GAN-derived models were developed for image generation and translation tasks \cite{2019Recent}. A typical GAN comprises a generator and discriminator that compete with each other to attain Nash equilibrium. The purpose of the generator is to produce as much synthetic data as possible that aligns with the potential distribution of real data, whereas the discriminator's aim is to accurately differentiate between genuine and fabricated data. The architecture of the GAN prototype is illustrated in \textbf{Fig}.\ref{fig:1}. The input of the generator is random noise, denoted by $z$, which is mapped into a new data space using function G($z$). The discriminator serves as a binary classifier that differentiates between real samples taken from the dataset and fake samples generated by the generator. During adversarial training, the objective function aims to maximize generator loss and minimize discriminator loss. When the discriminator cannot distinguish between real and fake data, it reaches an optimal state. At this point, the generator successfully learns the distribution of the real data.
\indent Although GANs have achieved significant success in 2D image synthesis, the generated images cannot preserve the 3D consistency. In contrast, the neural radiation field (NERF) \cite{2020NeRF}, which is briefly summarized as the use of an MLP network to learn a 3D geometric representation from a set of posed images, enables the rendering of images from an arbitrary view because it is a continuous 3D presentation of 2D images with camera poses. Due to the inherent features of the radiance fields, rendered images can enforce multiview consistency. Currently, neural radiation fields have been successful in applications of reverse rendering, novel view synthesis, 3D object editing, digital human bodies, and image/video processing. The primary limitations of NERFs are that they require posed images for training and are unable to learn multiple scenes using a single MLP.\\
\indent By integrating a neural radiance field into the generator, a generative radiance field (GRAF) \cite{2020GRAF} was implemented to produce 3D-aware images from random noise with a Gaussian distribution, and the model was trained on unstructured datasets without 3D supervision. The conditional radiance field in the generator uses 5D coordinates with the spatial location $(x, y, z)$ and viewing direction $(\theta,\phi)$ as inputs, and novel views are synthesized by projecting the output color $c$ and density $\theta$ into an image using differential volume rendering. A patch-based discriminator was employed to distinguish between fake and real images. Furthermore, the shape and appearance are modeled by $z_{s}$ and $z_{a}$ respectively, to manipulate them separately during the inference. The shape variable $z_{s}$ and the appearance variable $z_{a}$ were obtained separately by sampling a Gaussian distribution. \\
\indent GRAF is capable of disentangling shapes from appearance using shape and appearance codes and taking precise control of the camera pose for novel view synthesis, and does not require posed images for training. However, a bottleneck is that one MLP merely represents a scene, resulting in high memory overhead in the case of multiple scenes. Furthermore, GRAF cannot provide sophisticated control over the shape and appearance of generated objects. To address these issues, we introduce \textbf{CtrlNeRF}, a generative model based on neural radiance fields that allows precise control of image synthesis according to class labels (shape) and color labels (appearance). The framework is illustrated in \textbf{Fig}.\ref{fig:2}. The generated images preserved 3D consistency because of the intrinsic features of the radiance field in the generator. Moreover, the model allows for novel view synthesis by manipulating the camera pose. Specifically, we make the following contributions.
\begin{itemize} 
\item We modified the input and output of the MLP and added a VGG-based discriminator to differentiate class and color.
\item We represented multiple scenes using a single MLP, reducing storage consumption and increasing inference efficiency.
\item We achieved explicit control over the 3D-aware image generation according to the class and color labels.
\item We synthesized novel views nonexistent in the dataset through camera pose alteration and feature interpolation.\\
\end{itemize}

\indent The remainder of this paper is organized as follows. Section II introduces related work on 2D/3D image generation. Section III briefly reviews backbone models and explains the proposed method. Section IV presents the experimental settings and evaluation metrics. Section V presents a qualitative and quantitative analysis of the results. Finally, Section VI concludes the paper. 
\section{Related Works}
\textbf{2D Image Synthesis:} Generative adversarial networks (GANs) are deep generative models that perform advanced unsupervised tasks \cite{2017Asurvey} such as image generation, image superresolution, and text-to-image synthesis, $etc$. For unconditional GANs, for example DCGAN \cite{2017Unsupervised}, the input of the generator is random noise, which is an unrestricted input that probably leads to low quality images in some cases, and the formation of an image is uncontrollable due to the randomness of the input. In contrast, conditional GANs, such as InfoGAN \cite{2016InfoGAN}, CGAN \cite{2014Conditional}, and ACGAN\cite{2016Conditional}, incorporate conditional variables (labels and text) into the generator and discriminator, allowing the generation of high-quality images with control. To stabilize the training process, the Wasserstein generative adversarial networks \cite{2017Wasserstein}\cite{2017ImprovedWasserstein}\cite{2017OnWasserstein} used the Earth-Mover (EM) distance to optimize the objective function, producing a better gradient behavior than other distance metrics.

For a typical GAN, obtaining high-resolution images is challenging because the discriminator can easily distinguish between false and true images at high resolution. Several strategies have been implemented to enhance the stability of the training process and progressively improve image resolution \cite{2017Progressive} \cite{2017StackGAN} \cite{2017StackGAN++}. For example, a GAN HD pixel-to-pixel \cite{2017High} can produce high-resolution images up to 2048 × 2048 pixels. To improve image generation control, several studies \cite{2019A}\cite{2020High}\cite{2020Semi}\cite{2014Learning} have been conducted to disentangle the underlying factors of variation. Two-dimensional images are essentially projections of three-dimensional objects. However, they cannot ensure multiview consistency owing to the absence of 3D geometric constraints.\\

\textbf{Implicit Representation:} Implicit representations of 3D geometry are popular for deep learning 3D reconstruction \cite{2021Advances}. The advantages of voxel-based \cite{2016Generative}\cite{2016Unsupervised} \cite{2017OctNet}\cite{2016Learning} or mesh-based methods \cite{2018AtlasNet}\cite{2018Deep}\cite{2020Deep}\cite{2018Pixel2Mesh} are that implicit representations are continuous and are not restricted to topology. Recently, hybrid grid representations \cite{2020Local}\cite{2020Convolutional} have been extended to large-scale scenes, but all of the above methods require 3D input without considering texture. To overcome the limitations of 3D supervision, some studies \cite{2019DIST}\cite{2020Differentiable}\cite{2019Scene}\cite{2020Pix2Vox} presented differentiable rendering techniques to learn continuous shape and texture representations from 2D-posed images. Mildenhall et al. \cite{2020NeRF} proposed neural radiance fields, in which they combined an implicit neural model with volume rendering for novel view synthesis. NeRF requires multi-view images with camera poses for supervision and trains a single network per scene. The neural radiance field is implemented by an MLP, which is a fully connected multilayer network with a 5D input of spatial location $\textbf{x} (x,y,z)\in\mathbb{R}^{3}$ and viewing direction $\textbf{d} (\theta, \phi)\in\mathbb{R}^{2}$, 4D output of volume density $\sigma\in\mathbb{R}^{+}$ and view-dependent color $\textbf{c} (r,g,b)\in\mathbb{R}^{3}$. 

\textbf{3D-Aware Image Generation:} To date, neural scene representations have been integrated into generative models to enable the synthesis of 3D-aware images from latent code. Voxel-based GANs \cite{2019Escaping}\cite{2019HoloGAN}\cite{2020BlockGAN} learn textured 3D voxel representations from two-dimensional images using differentiable rendering techniques. However, such voxel-based models are memory intensive, impeding high-resolution image synthesis. Radiance field-based methods \cite{2021pi} achieve higher quality and better 3D consistency, but have difficulties in training high-fidelity images due to the cost of the rendering process. 

Mildenhall et al. \cite{2020NeRF} proposed neural radiance fields (NERF) that can implicitly represent 3D geometries and synthesize novel views using volume rendering.  NERF and their variants are valuable tools for generating 3D-aware images. Despite their strengths, they are limited by slow training and inference, inability to handle dynamic scenes, generalization shortcomings, and the necessity for a great number of perspectives. To address these challenges, Garbin et al. \cite{2021FastNeRF} introduced FastNeRF, a method that can generate high-quality images at a rate of up to 200 Hz. To apply NERF to unknown scenes, studies on this issue include pixelNeRF \cite{2020pixelNeRF} and IBRNet\cite{2021IBRNet}. Furthermore, J. Gu et al. \cite{2021StyleNeRF} proposed styleNeRF to synthesize high-resolution images at interactive rates, allowing control of camera poses and different levels of styles. Huang et al. \cite{2022StylizedNeRF} designed a framework for stylizing 3D scenes through 2D-3D mutual learning.

Taking advantage of both GAN and NeRF, Schwarz \emph{et al.} \cite{2020GRAF} introduced generative neural radiance fields (GRAF). Although the shape and appearance are disentangled in the model, they are restricted to a single-object scene without explicit control of the image synthesis. Niemeyer \emph{et al}. \cite{2021GIRAFFE} presented GIRAFFE to learn 3D representation of a compositional scene as synthetic neural feature fields, which employs MLP to represent each object in the scene and reconstruct them afterwards, significantly increasing memory consumption and computational cost. Most recently, several SOTA generative models inspired by GRAF have been introduced. HeadNeRF \cite{2021HeadNeRF} is a facial rendering method that combines NeRF and facial parameterization models. Its outstanding advantages lie in real-time performance and support for separate control of camera pose, facial identity, expression, and appearance. GRAM \cite{2022GRAM} is an innovative method designed to control point sampling and learning of radiance fields on 2D manifolds, represented as a collection of implicit surfaces within a 3D volume. Clip-NeRF \cite{2021CLIP} is a versatile framework that enables intuitive manipulation of NeRF through brief text prompts or exemplar images. It combines NeRF's capability for novel view synthesis with the controllable manipulation of latent representations in generative models.
\section{Method}
The neural radiance field (NERF) has achieved impressive results in a novel view synthesis using a set of posed images. Combined with the generative model, the generative radiance field (GRAF) has been successfully employed in 3D-aware image synthesis from latent code. The generated images preserve multiview consistency due to the benefits of the neural radiance field. Moreover, the GRAF prototype can be trained using unposed images and provides explicit control over the camera pose. The shapes and appearances in GRAF were disentangled using the shape code $z_a$ and the appearance code $z_s$.  However,  shapes and appearances are subject to a certain level of unpredictability because of the randomness of latent codes. Our approach employs a single MLP to learn multiple scenes and achieves precise control over the synthesis of 3D images based on labels. To support the rationale behind our model design, we initially present the fundamentals of NERF and GRAF.\\

\textbf{Neural Radiance Fields (NERF)}: The radiance field is a continuous representation of a scene, denoted by the function $F_{\Theta}$, which takes the 3D location \textbf{x}, the viewing direction \textbf{d} as input and the color \textbf{c} along with the volume density values $\sigma$ as output. The mapping function $F_{\Theta}$:($\textbf{x}, \textbf{d}$) $\rightarrow$(\textbf{c}, $\sigma$) was implemented using a fully connected network that optimizes the weights to map each of the 5D coordinate inputs to their appropriate density and color. Due to the bias of deep networks towards lower frequency functions, the function $F_{\Theta}$, when applied directly to the 5D coordinate input, proved inadequate to capture high frequency variations. Hence, positional encoding $\gamma$() is used to translate a 3D location and viewing direction into a high-dimensional space, thus facilitating $F_{\Theta}$ to approach a high-frequency function with greater ease, formally defined in Equation \ref{eq:eq1}.
\begin{equation}
\begin{aligned}
\gamma(p)= &[sin(2^0\pi p), cos(2^0\pi p), (sin(2^1\pi p), cos(2^1\pi p), \\
&\dots, (sin(2^{L-1}\pi p), cos(2^{L-1}\pi p)]
\end{aligned}
\label{eq:eq1}
\end{equation}
\\ The function $\gamma$() is applied independently to the three coordinate values (x,y,z) of the position \textbf{x} and two components of the unit vector of the viewing direction \textbf{ d}. The MLP network assigns the resulting characteristics to the color value $c\in\mathbb{R}^3$ and the volume density $\sigma\in\mathbb{R}^{+}$, as shown in Equation \ref{eq:eq2}. Here, $L_{x}$=10 and $L_{d}$=4.
\begin{equation}
\begin{aligned}
\quad\quad\quad\gamma(\textbf{x}),\gamma(\textbf{d})&\mapsto(\textbf{c},\sigma)\\
\quad\quad\quad\mathbb{R}^{L_x} \times \mathbb{R}^{L_d}&\rightarrow\mathbb{R}^3\times\mathbb{R}^+\\
\end{aligned}
\label{eq:eq2}
\end{equation}
\indent The neural radiance field is a representation of a scene as the volume density and emitted radiance at every point in space. The volume density, denoted by $\sigma$, can be thought of as the probability differential of a ray terminating at an infinitesimal particle at a specific location \textbf{x}. The expected color $C(r)$ of the camera ray: $r(t) = o + td$ is defined in Equation \ref{eq:eq3}, with near and far bounds $t_n$ and $t_f$.
\begin{equation}
\begin{aligned}
&C(r) =\int_{t_n}^{t_f} T(t)\sigma(r(t))c(r (t),d))dt\\
&where \quad T(t) =\exp{(-\int_{t_n}^{t}\sigma(r(s))ds)}
\end{aligned}
\label{eq:eq3}
\end{equation}
where the function $T(t)$ denotes the accumulated transmittance along the ray from $t_n$ to $t$ and the probability that the ray travels from $t_n$ to $t$ without hitting any other particles. Rendering a 2D image from a neural radiance field requires estimating the integral $C(r)$ for each camera ray $r$ traced through the pixels of a virtual camera.\\
\indent The integral of $C(r)$ is typically estimated using a deterministic quadrature, which inherently restricts the resolution of rendered images because the MLP is merely interrogated at a discrete points. A stratified sampling approach was implemented to divide the data into uniformly spaced intervals, from which a single representative sample was randomly selected from each interval. This method computes the integral value by aggregating data points from a discrete collection of samples. Moreover, a hierarchical volume sampling technique was used to enhance rendering efficiency.\\

\textbf{Generative Radiance Fields (GRAF):} Generative Radiance Field (GRAF) is a generative model comprising a generator based on the radiance field and a multi-scale patch discriminator. This model enables the synthesis of 3D-aware images from random noise, and is trained on unposed datasets.\\

(1) \textbf{Generator:} The inputs of the generator is made up of the intrinsic camera parameter K, the camera pose $\xi$, the sampling pattern $\nu$, the shape code $z\_s$, and the appearance code $z_a$. The generator generates predicted image patches, denoted as $P^{'}$, as its output. The pose of the camera, denoted $\xi$, is randomly selected from the pose distribution, denoted by $p_\xi$. The center $(u,s)$ and scale of the virtual patches are determined using a uniform distribution. Furthermore, the shape and appearance codes, denoted by $z_s$ and $z_a$, are drawn from the shape and appearance distributions, denoted by $p_a$ and $p_s$.\\ 
\textbf{\emph{Ray Sampling}}:The real patch $P(u,s)$ is determined by utilizing the 2D image coordinates that specify the position of each pixel in the image domain. The corresponding rays are determined by these coordinates, the intrinsic camera parameter K, and the camera pose $\xi$.\\\\
\textbf{\emph{3D Point Sampling}}: The sampling method involves sampling N points $\{\textbf{x}^i_r\}^N_{i=1}$ along each ray $r$ for the numerical integration of the expected color $C(r)$. Instead of using a single network to represent the scene, stratified sampling optimizes two networks: one 'coarse' and one 'fine' simultaneously. This procedure allocates more samples to the visible region to increase the quality of the images.\\
\textbf{\emph{Conditional Radiance Field}}: The conditional radiance field is implemented by a fully connected neural network with parameter $\theta$. More than a regular radiance field, it is subject to the inputs of shape code $z_s$ and appearance $z_a$. The encoding for shape $h$ is obtained by concatenating  the positional encoding $\gamma(x)$ and shape code $z_s$ and is subsequently converted to the volume density $\sigma$ through a density head $\sigma_\theta$. Nevertheless, in order to separate the shape and appearance, the volume density was independently predicted without employing the view direction $d$ and appearance code $z_a$  during the inference process. To estimate the predicted color $\textbf{c}$, a concatenating vector comprising the shape encoding $h$, positional encoding of the direction $\gamma(d)$, and appearance code $z_a$ is fed into the color head $c_\theta$ for further inference. \\
\textbf{\emph{Volume Rendering}}: The acquisition of the color $\textbf{c}$ and volume density $\sigma$ of N points along the ray $(c_r^i,\sigma_r^i)$ was achieved using volume rendering. The synthesized patch $P^{'}$ was obtained by combining the result of every sampling ray, and the value of the color $c_r$ was calculated using equation \ref{eq:eq5}.
\begin{equation}
\begin{aligned}
c_r=\sum_{i=1}^N T_r^i\alpha_r^i&c_r^i\quad\quad   
T_r^i=\prod_{j=1}^{i-1}(1-\alpha_r^j)\\
\alpha_r^i=&1-\exp{(-\sigma_r^i\delta_r^i)}
\end{aligned}
\label{eq:eq5}
\end{equation}
The transmittance ($T_r^i$) and alpha value ($\alpha_r^i$) of sample point $i$ along the ray r are denoted by $T_r^i$ and $\alpha_r^i$, respectively, and the distance between neighboring sample points is defined by $\delta_r^i=\left\|x_r^{i+1}-x_r^{i}\right\|_2$.\\

(2) \textbf{Discriminator:} The development of a discriminator involves the construction of a deep convolutional neural network (CNN) with ReLU as an activation function. The discriminator accelerates both training and inference by comparing the synthesized patch $P^{'}$ with the real patch $P$, which is obtained by accessing a real image at 2D coordinates $P(u, s)$ through bilinear interpolation, referred to as $\Gamma(I,\nu)$. The discriminator was adequate for all patches randomly sampled on various scales. The size of the patch determines its receptive field, where larger receptive fields are utilized to capture global content, and smaller receptive fields are used to progressively discern local details.\\

(3) \textbf{ Training and inference:}  In adversarial training, the generator $G(\theta)$ seeks to minimize the function $V(\theta,\phi)$, while the discriminator $D(\phi)$ seeks to maximize it. The nonsaturating objective function $V(\theta,\phi)$ with R1 regularization is defined in Equation \ref{eq:eq10}.
\begin{equation}
\begin{aligned}
&V(\theta,\phi)=\mathbb{E}_{z_s\sim p_s, z_a\sim p_a, \xi\sim p_\xi, \nu\sim p_\nu}[f(D_\phi(G_\theta(z_s ,z_a,\xi,\nu)))]\\
&+\mathbb{E}_{I\sim p_D,\nu\sim p_\nu}[f(-D_\phi(\Gamma(I,\nu)))-\lambda\|\nabla D_\phi(\Gamma(I,\nu))\|^2]
\end{aligned}
\label{eq:eq10}
\end{equation}
Where $f(t) = -\log(1 + \exp(-t))$, $I$ signifies an image sampled from the data distribution $p_D$, and $p_\nu$ refers to the distribution over random patches. Furthermore, the parameter $\lambda$ controls the level of regularization. The discriminator utilizes both spectral normalization and instance normalization.\\
\\
\textbf{CtrlNeRF:} Actually, GRAF can generate a 3D geometry from latent code; however, its shape and appearance are not easily manipulated. To address this issue, we developed a GRAF-derived model (i.e. \textbf{CtrlNeRF}). This model allows us to use a single MLP to learn multiple 3D representations and to explicitly control object formation. We modified the output of the MLP to disentangle the shape and appearance and added an extra discriminator to distinguish between the object category and style.\\

\begin{figure}[htbp]
\centering
\includegraphics[width=7cm]{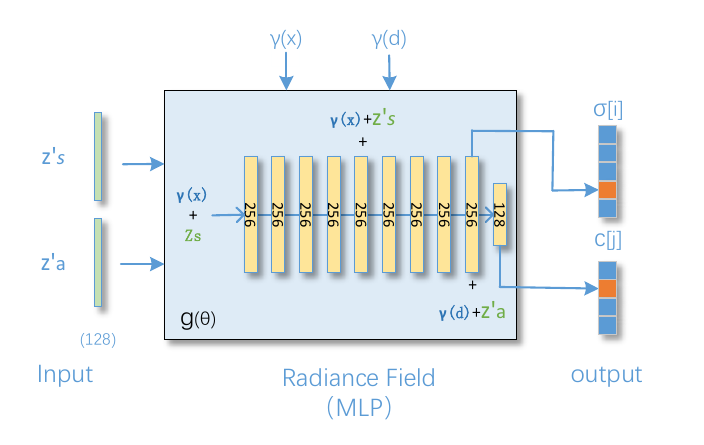}
\caption{The architecture of a conditional radiance field (MLP) comprises inputs of $z^{'}_{s}$ and $z^{'}_{a}$, as well as $\gamma(x)$ and $\gamma(d)$. The output of the model consists of a volume density array $\sigma$[] and color array c[].}
\label{fig:6}
\end{figure}

(1) \textbf{Generator:} Based on the GRAF generator, we manipulated the input of the MLP by embedding label codes in shape and appearance codes, allowing the generator to utilize the label-embedded codes to generate a 3D geometry with precise controls of shape and color.
\begin{figure}[htbp]
\centering
\includegraphics[width=7cm]{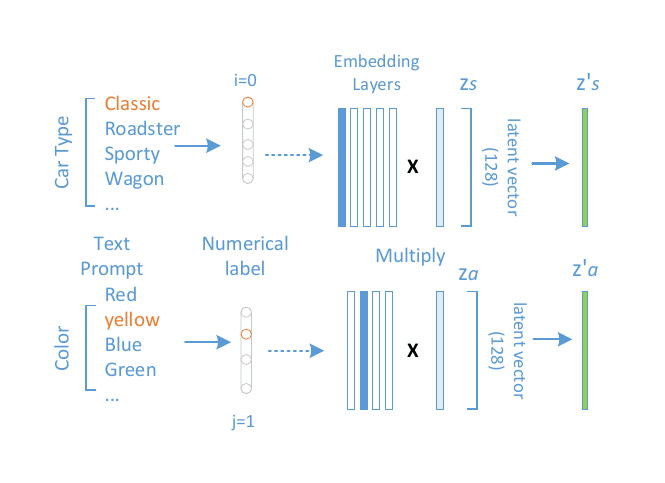}
\caption{Scheme for incorporating label codes into a latent code through multiplication.}
\label{fig:5}
\end{figure}\\
\textbf{\emph{Input}}: The latent code $z$ is typically separated into two components: shape code $z_s$ and appearance code $z_a$. The text labels were initially translated into numerical labels (i=0, 1, 2, ... ), and the corresponding vector was extracted from the embedding layers using the label index as a reference. The label-embedded codes $z^{'}_{s}$ and $z^{'}_{a}$ were derived by multiplying the latent codes $z_{s}$ and $z_{a}$ by the feature vectors.\\
\textbf{\emph{Output}}: Unlike the GRAF, the output of MLP in the model consists of a volume density array $\left[\sigma(i)\right]^{N-1}_{i=0}$ and a color array $\left[c(j)\right]^{M-1}_{j=0}$. In this context, M denotes the number of classes, and N denotes the number of styles. The label i/j refers to the numerical representation of the class/color. The label embedding technique is illustrated in Fig.\ref{fig:6}.
\\
\textbf{\emph{Conditional Radiance Field}}: The inference for the volume density and color resembles that of the GRAF prototype. However, the MLP in the generator is conditional on the inputs of the label-embedding latent code, and the outputs are the density and color arrays associated with class and style. The structure of the proposed conditional radiance field is depicted in \textbf{Fig}.\ref{fig:5}. In this design, $\gamma(\textbf{x})$ and $\gamma(\textbf{d})$ refer to the positional encoding for the coordinates $\textbf{x}$ in the 3D space and the directions $\textbf{d}$ of the rays associated with each point, respectively. $z^{'}_{s}$ and $z^{'}_{a}$ are the label-embedded latent codes. \\

(2) \textbf{Discriminator}: In addition to using the typical discriminator in GRAF to evaluate the generated patch $P^{'}$ compared to the real patch $P$,  we employed a discriminator based on VGG16 \cite{2015Very} to effectively classify various classes and styles of objects. The VGG network is well known for its exceptional performance in multi-classification tasks. Initially, the discriminator $D_{vgg}$ was trained using annotated images $I^{'}$ that were down-scaled from the real image $I$. The pre-trained discriminator was utilized as an auxiliary classifier for the patches generated $P^{'}$.  To further improve image quality, we adopted posed images for training and replaced adversarial loss with reconstruction loss.\\

(3) \textbf{ Training and Inference}: During supervised learning, the network parameters are optimized using loss functions. In particular, the discriminator was trained by a real patch $P$ and a generated patch $P^{'}$ to improve computational efficiency. The discriminator $D_{vgg}$ was trained in real images resized $I^{'}$. The loss function to update the weights of the network is defined in Equation \ref{eq:eq7}, and the pseudocode for training is shown in \textbf{Algorithm} \ref{alg:euclid}.
\begin{equation}
\begin{aligned}
L(G(\theta))=& L_{adv} (D(\phi)|P^{'}_{i,j}) + \lambda_{1} L_{cls} (D_{vgg}|P^{'}_{i,j})\\&+\lambda_{2} L_{sty}(D_{vgg}|P^{'}_{i,j})
\end{aligned}
\label{eq:eq7}
\end{equation}
where, $L_{adv}$ refers to adversarial loss between $P$ and $P^{'}$, $L_{cls}$ and $L_{sty}$ refer to  the loss of class and  style, respectively, and $\lambda_{1}$ and $\lambda_{2}$ denotes weights for $L_{cls}$ and $L_{col}$.  In the experiment, RMSprop was used as optimizer and the weights of these losses were $\lambda_{1}$=2.0, $\lambda_{2}$=3.0, with a batch size of 8.
\begin{algorithm}[ht!]
\caption{CtrlNeRF training algorithm}
\label{alg:euclid} 
\renewcommand{\algorithmicrequire}{\textbf{Input:}}
\renewcommand{\algorithmicensure}{\textbf{Initialization:}}
\begin{algorithmic}
\Require  real images $I$ with labels $(\hat{i},\hat{j}).$ 
\Ensure  camera intrinsic K, camera pose  $\xi$, and sampling pattern $\nu$.
\item 
$\textbf{do}\ \ iterations$\\
$\quad random\ \ (i,j)\in (\hat{i},\hat{j})$ \\
$\quad z^{'}_{s},z^{'}_{a} \gets z_{s}, z_{a}$\\

$\quad P^{'}_{i,j} \gets G(z^{'}_{s},z^{'}_{a}):$\\
\quad\quad\quad\quad \textbf{for} $\ M \ points \ along \ N \ rays:$\\
\quad\quad\quad\quad$\sigma[i], \textbf{c}[j] \gets F_{\Theta}(\textbf{x},\textbf{d})$)\\
\quad\quad\quad\quad$P^{'}_{i,j} \gets \pi(\sigma[i], \textbf{c}[j])$\\
\quad\quad\quad\quad \textbf{end \ for} \\
$\quad P_{\hat{i},\hat{j}} \gets \mathcal{T}(I)$ \\
$\quad L_{adv} \gets D_{\phi}(P^{'}_{i,j},P_{\hat{i},\hat{j}})$ \\ 
$\quad L_{cls},L_{sty} \gets D_{vgg}(P^{'}_{i,j}, I)$ \\ 
$\quad Loss= L_{adv}+\lambda_{1}L_{cls}+\lambda_{2}L_{sty}$\\
$\quad update \ \ G$\\
$\quad update \ \ D_{\phi},D_{vgg}$\\
{\textbf{end do}}
\end{algorithmic}
\end{algorithm}
\section{Experiments}

\subsection{Datasets}
In this study, due to the lack of annotations in the datasets used for generative models such as GRAF and GIRAFFE, we began by developing a synthetic dataset named \textbf{CARs} (I) utilizing 3D editing software. First, a car was situated at the origin of the coordinate system, and a virtual camera was placed on the surface of the upper hemisphere oriented towards the origin. where $\theta$ and $\phi$ represent the pitch and yaw angles of the camera, respectively. The camera was placed in a hemisphere with a radius of $r$. By manipulating the pose of the virtual camera, we could obtain the views of an object with variable respect. The captured images with a size of 800x800 were automatically labeled by class, color, and pose. Four types of cars (classic, roadster, sporty, and wagon) were included in the CARs dataset, each of which was presented in four color modes: red, green, blue, and yellow. In addition, we used publicly accessible NERF datasets \cite{2020NeRF}, specifically  \textbf{Synthetic} (II) and \textbf{LLFF} (III), for demonstration.
\subsection{Baselines}
To demonstrate its excellence in 3D-aware image generation, we compared our model with the latest NeRF-based generative models, including CLIP-NeRF. NeRF can learn the continuous representation of 3D geometry from posed images using neural radiance fields and render a novel view using differentiable volumetric rendering. GRAF \cite{2020GRAF} is a generative model capable of producing images with 3D consistency using latent codes related to shape and appearance, without requiring 3D supervision. GRIFFEE \cite{2021GIRAFFE} is another generative model derived from the GRAF model, which uses multiple MLPs to represent compositional scenes. CLIP-NeRF is the SOTA multimodal 3D object manipulation method for neural radiance fields using a short text prompt. For these generative models, qualitative and quantitative analyzes were performed to evaluate the performance of the proposed model by comparing it with other generative models.

\subsection{Evaluation Metrics}
The FID score, which comprises human assessments of realism and diversity, has been widely used to evaluate the quality and variety of the generated images. This metric was first introduced by Kanazawa et al. in 2018 \cite{2018An}. The lower the FID score, the better the model performance. FID score was derived from Equation \ref{eq:eq8}.
\begin{equation}
    \begin{aligned}
   \mathcal{FID} &=d^2 ((m_r ,C_r ),(m_g ,C_g ))\\
       &=||m_r - m_g || ^2_2+\mathcal{T}r(C_r+ C_g - 2(C_r C_g )^{1/2})
    \end{aligned}
    \label{eq:eq8}
\end{equation}
\indent The pair ($m_r$, $C_r$) corresponds to real images, while the pair ($m_g$, $C_g$) corresponds to generated images. In each pair, $m$ represents the mean and $C$ represents the covariance. The KID score \cite{2018Demystifying}, which is an unbiased estimate that does not require a normal distribution hypothesis, was introduced for image evaluation.\\
\indent In addition, to measure the reconstructed shapes and their closest shapes in the ground truth, the peak signal-to-noise ratio (PSNR) and the structural similarity (SSIM) \cite{2013Image} were used for a quantitative comparison between real and synthetic images.
\section{Results}
\subsection{Controllable 3D-aware Image Synthesis based on the Labels.}
\begin{figure*}[b]
\begin{minipage}{0.95\linewidth}
    \centering
    \includegraphics[width=12cm]{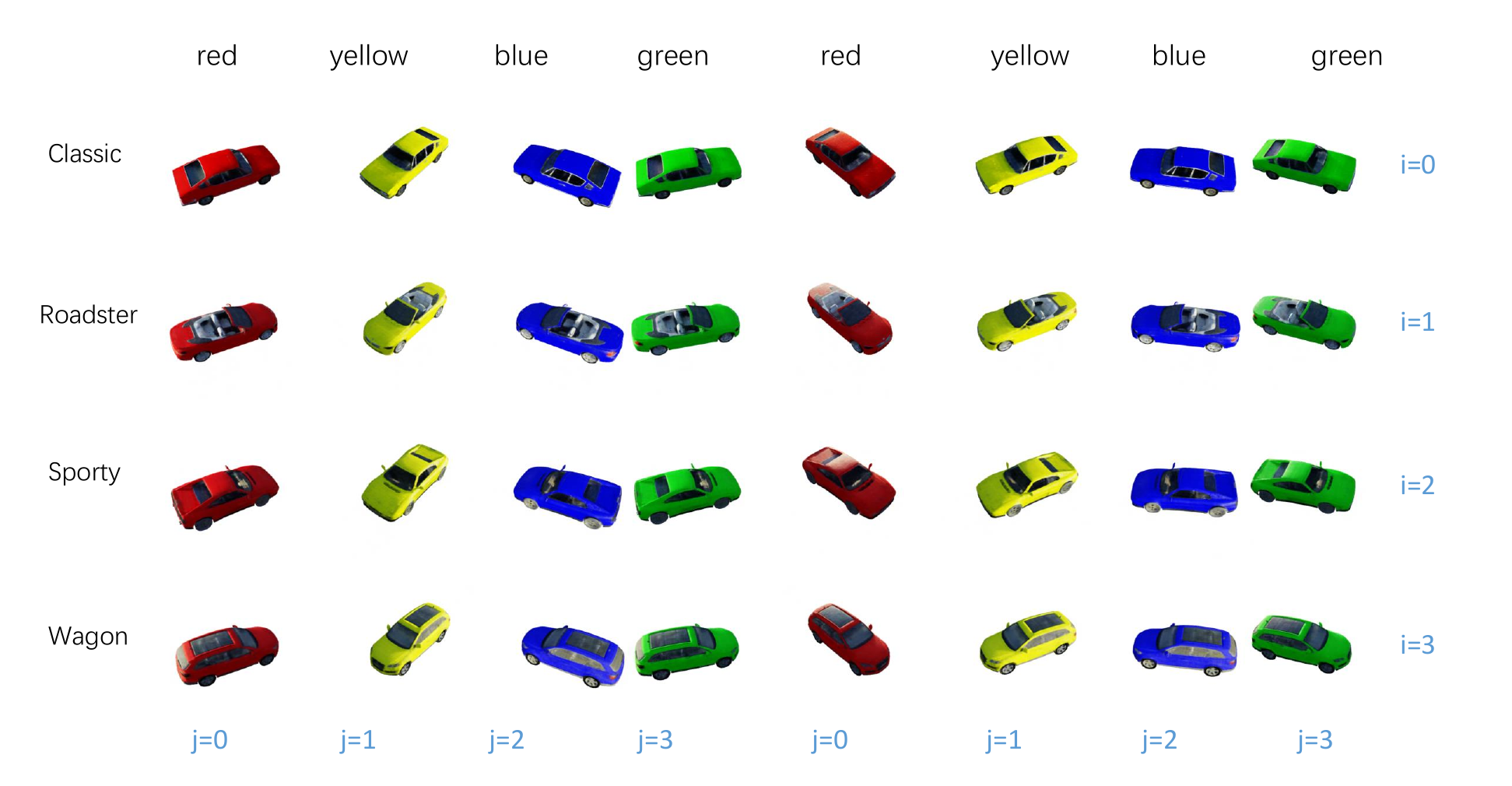}
    \caption{Samples of the synthesized images (400x400) on \textbf{CARs}(I) dataset.}
    \label{fig:7}
\end{minipage}
\begin{minipage}{0.95\linewidth}
    \centering
    \includegraphics[width=12cm]{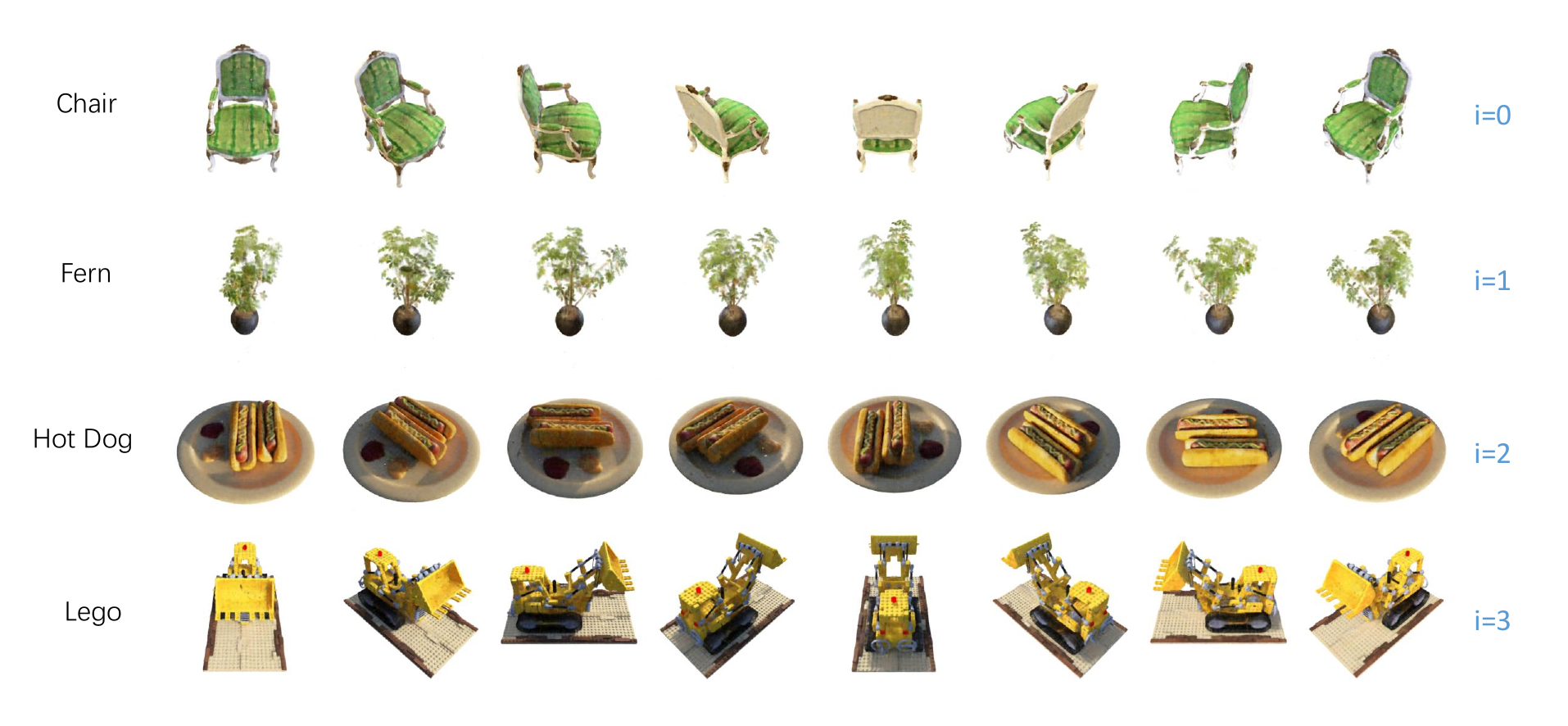}
    \caption{Samples of the synthesized images (400x400) on \textbf{Synthetic}(II) dataset.}
    \label{fig:8}
\end{minipage}
\begin{minipage}{0.95\linewidth}
    \centering
    \includegraphics[width=12cm]{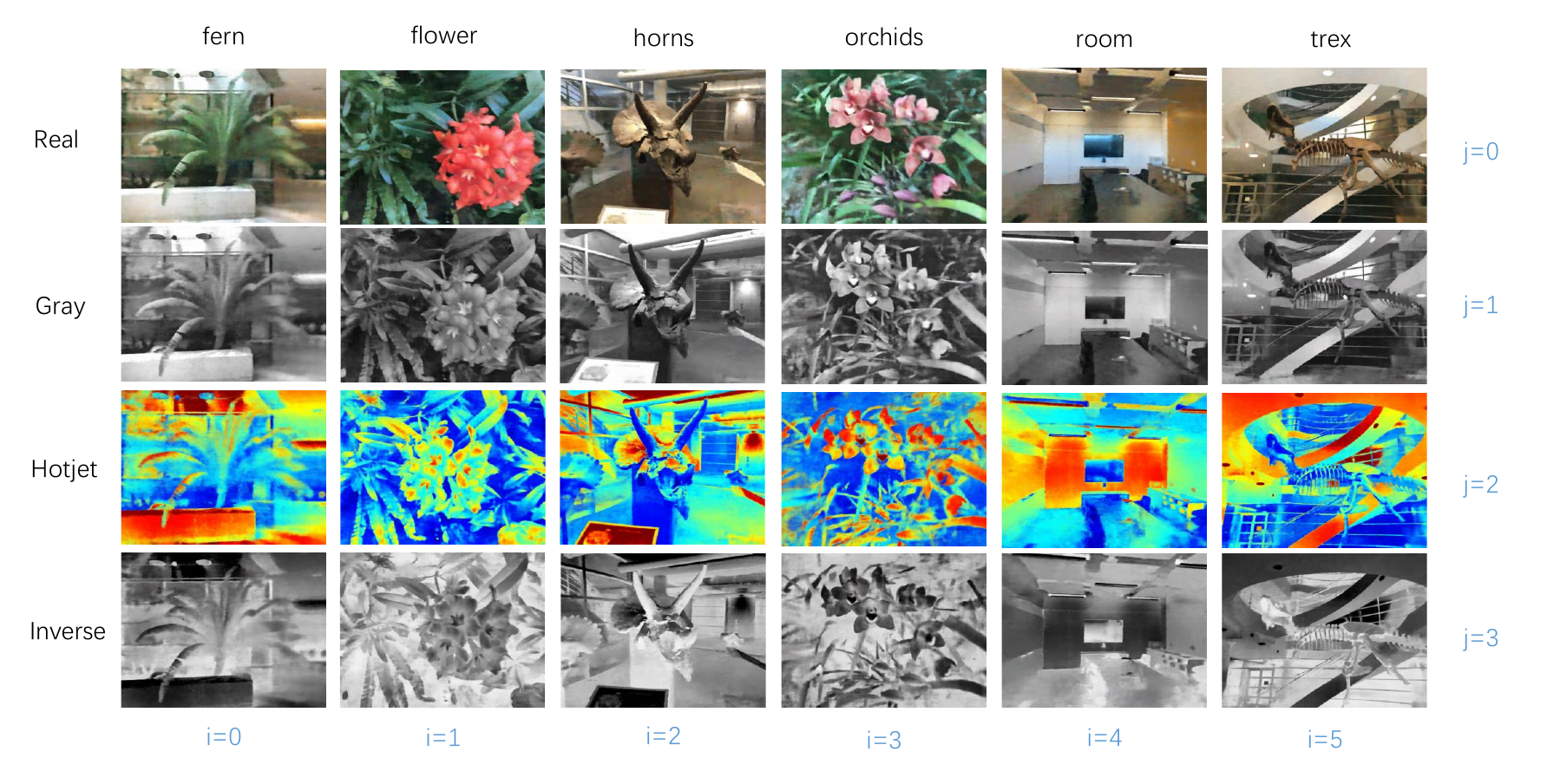}
    \caption{Samples of the synthesized images (504x378) on \textbf{LLFF}(III) dataset.}
    \label{fig:9}
\end{minipage}
\end{figure*}

\begin{figure}[htbp]
    \centering
    \includegraphics[width=6cm]{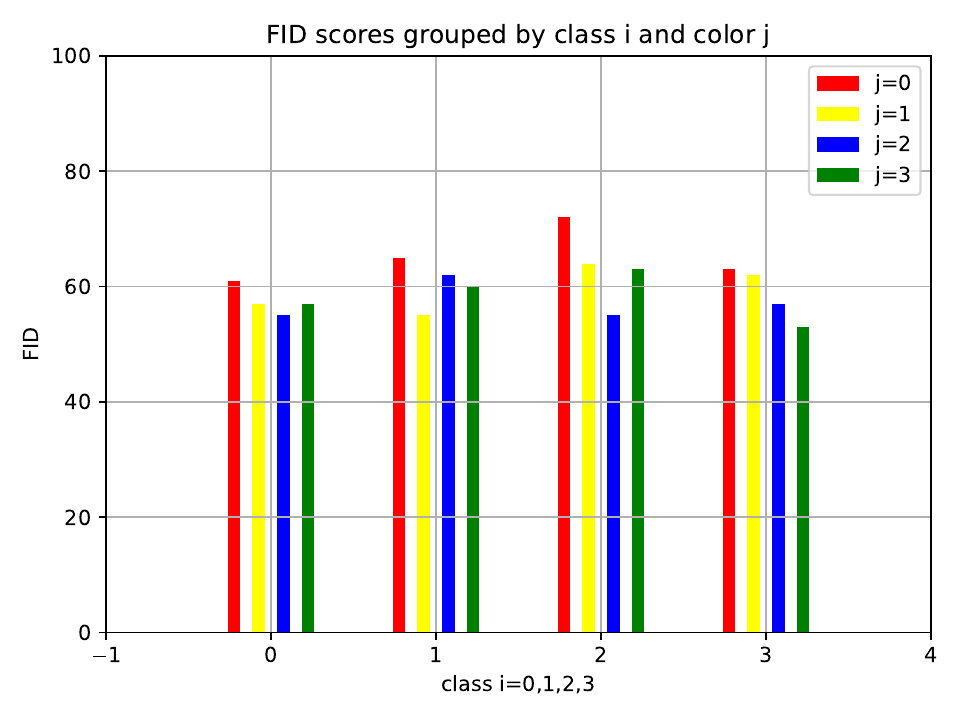}
\caption{The diagram of FID scores of the generated images grouped by class (i=0,1,2,3), and color (j=0,1,2,3) }
    \label{fig:10}
\end{figure}

\begin{figure}[htbp]
    \centering
    \includegraphics[width=7cm]{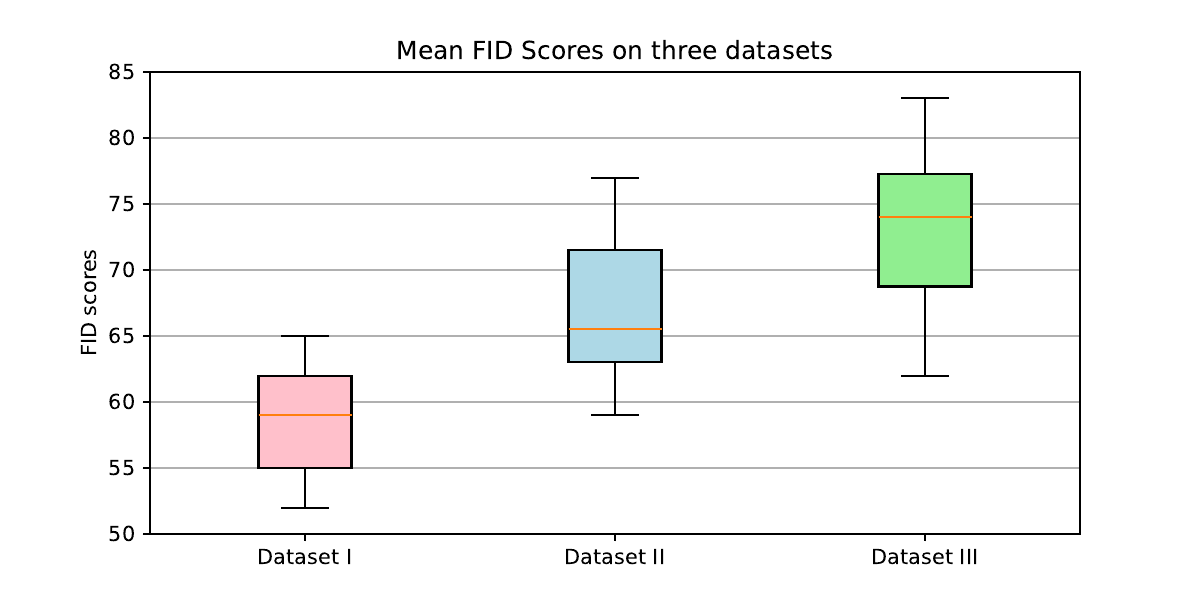}
\caption{The mean FID scores of the generated images on CARs(I), Synthetic(II) and LLFF(III) datasets, respectively}
    \label{fig:11}
\end{figure}

\begin{figure}[htbp]
    \centering
    \includegraphics[width=7cm]{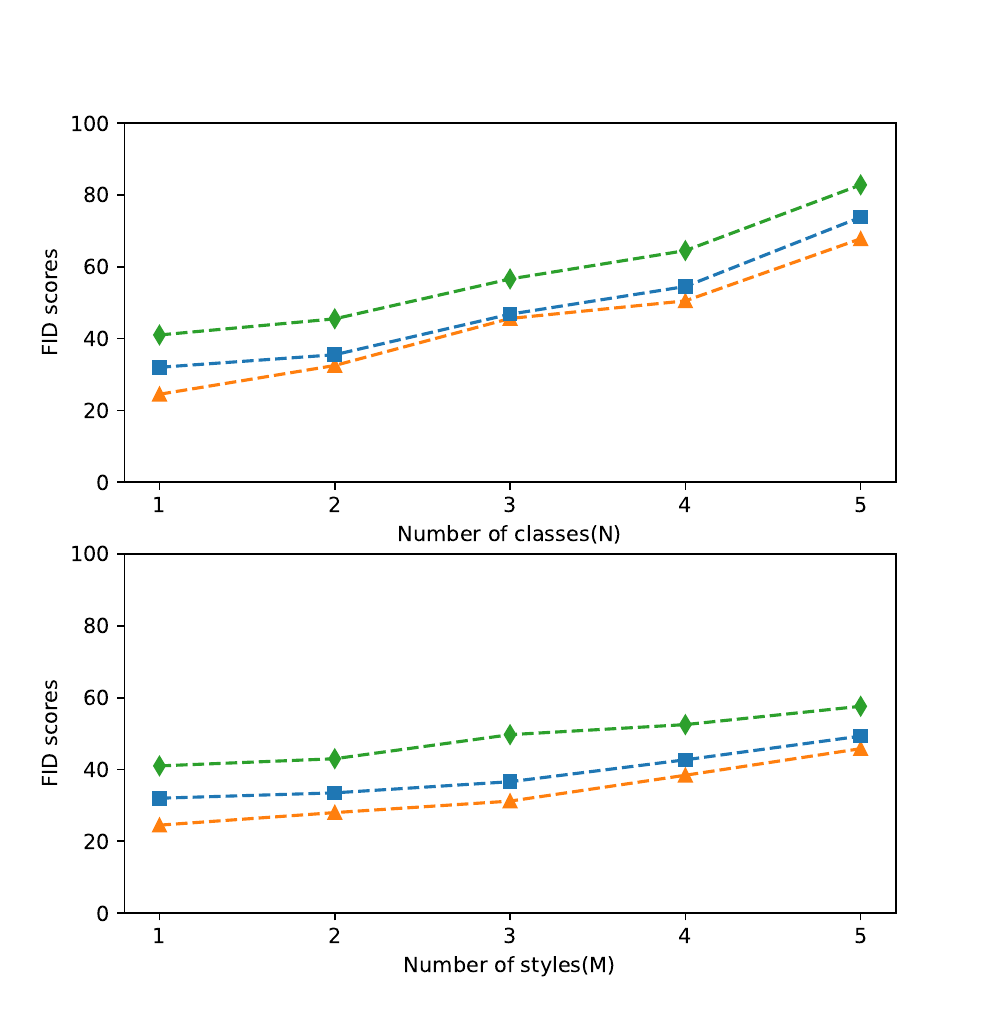}
\caption{The diagrams of FID scores of the generated images with increasing the number of classes and styles on CARs(orange), Synthetic(blue) and LLFF(green) datasets.}
    \label{fig:12}
\end{figure}

The model was trained on the \textbf{CARs}(I), \textbf{ Synthetic }(II), and \textbf{LLFF}(III) datasets. Label-associated latent codes $z^{'}_{s}$ and $z^{'}_{a}$ are the input of the MLP in the generator, and the view direction d$(\theta, \phi)$ is sampled within a specific pose range. In the experiment, we sampled 1024 rays for an image and 64 points along each ray. Therefore, 1024 × 64 points were sampled, each with 3D coordinates and ray directions. The RMSprop optimizer was used, with a learning rate of 0.0001 for the discriminator and 0.0005 for the generator, and class and color labels were used for the prediction. The synthesized images are shown in \textbf{Fig}.\ref{fig:7}, \ref{fig:8}, \ref{fig:9} for the three datasets. Optimization for multiple scenes typically requires approximately 100–200k iterations to converge on a single NVIDIA A4500 GPU(approximately 14 h). \\
\indent The FID score reflects the similarity and variance between real and synthesized images, which is an essential metric to quantitatively evaluate the performance of the generative model. The FID scores of the images generated in dataset I, grouped by class and color, are shown in \textbf{Fig}.\ref{fig:10}. When a single MLP handles multiple scenes within the model, it is evident that the image quality decreases as the number of scenes increases. The mean FID scores of the model trained on Datasets (I), (II), and (III) are presented in \textbf{Fig.}\ref{fig:11}. As shown in \textbf{Fig.} \ref{fig:12}. Image quality decreased with an increase in the number of classes and styles.

Furthermore, the model performance on the "LLFF" dataset was noticeably poorer compared to the "Car" dataset, likely due to the higher complexity of the images in the "LLFF" dataset, particularly when dealing with multiple scenes.  Higher complexity implies greater entanglement when using a shared-weight MLP to represent multiple scenes.
\subsection{Novel View Generation via Camera Manipulation.}
\begin{figure}[htbp]
    \centering
    \includegraphics[width=8cm]{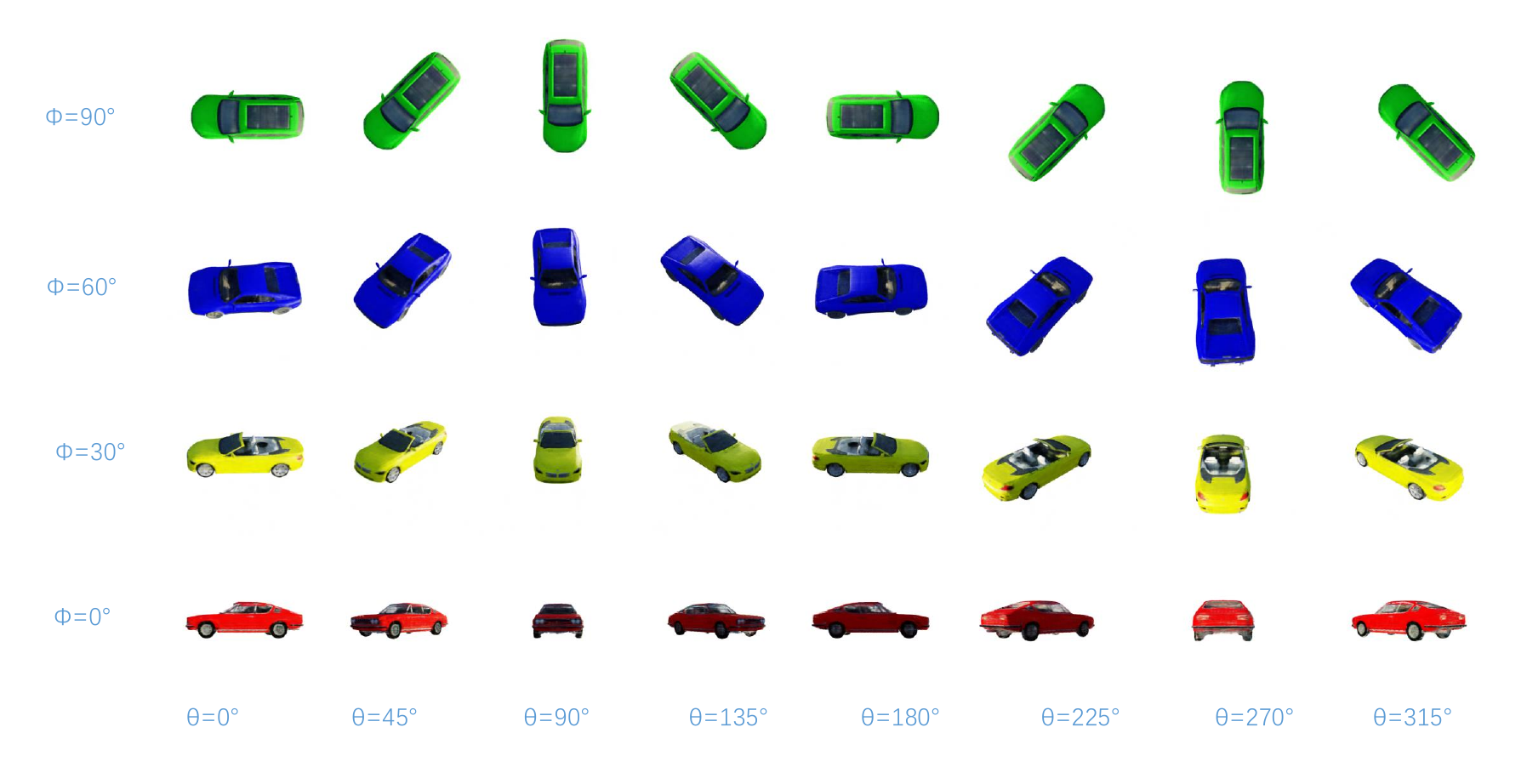}
    \caption{Novel views synthesized by manipulating the rendering camera. $\theta,\phi$ denotes the pitch or yaw angle, respectively.}
    \label{fig:13}
\end{figure}

\begin{figure}[htbp]
    \centering
    \includegraphics[width=8cm]{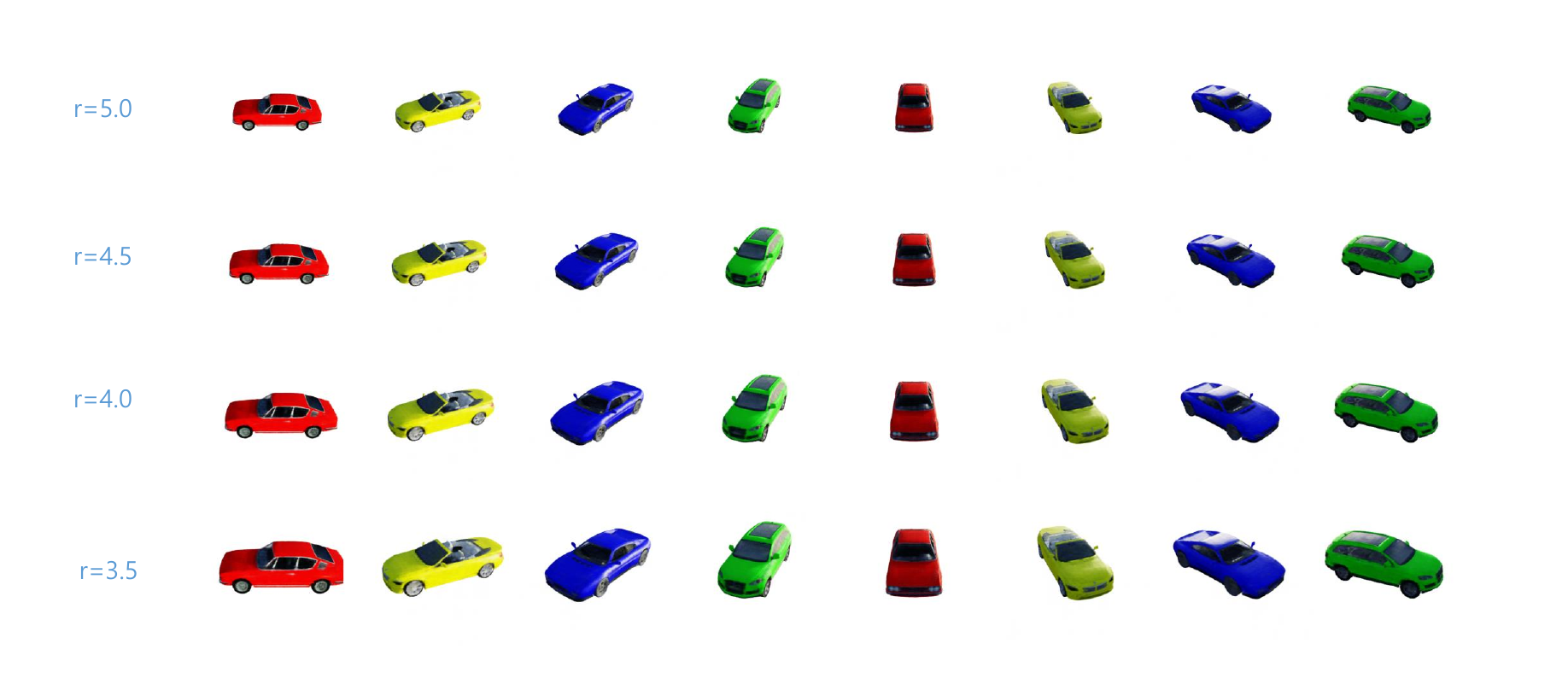}
    \caption{The depth translation of the synthetic object with the radius r from 3.5 to 5.0. Here, r is the distance between the origin of coordination and the camera.}
    \label{fig:14}
\end{figure}

\begin{figure}[htbp]
    \centering
    \includegraphics[width=8cm]{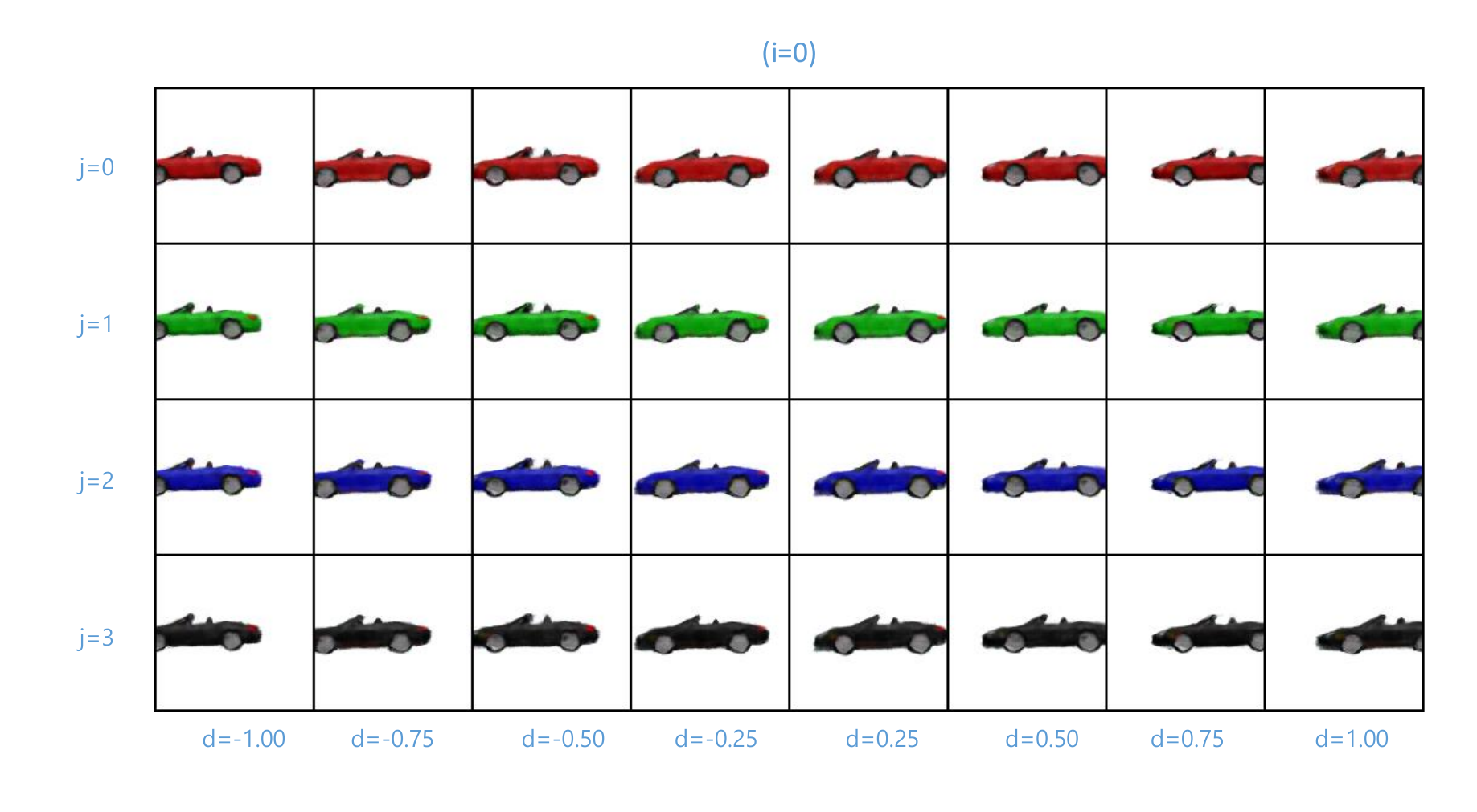}
    \caption{The horizontal translation of the synthetic object with the distance d from -1.0 to 1.0. Here, d is the horizontal shift of the camera.}
    \label{fig:15}
\end{figure}

As shown in the following three figures, novel views of an object can be obtained by alternating the poses of the rendering camera. For example, in \textbf{Fig.}\ref{fig:13}, the virtual camera captures images of the object with the poses of $\theta\in[-180^{\circ},180^{\circ}]$ and $\phi\in[0^{\circ},90^{\circ}]$. In \textbf{Fig.}\ref{fig:14}, we changed the radius of the rendering sphere stepwise, ranging from 3.5 to 5.0, with an interval of 0.5. Finally,  we performed horizontal translation of the synthesized objects within the range of (-1.0, 1.0) in \textbf{Fig.}\ref{fig:15}.
\subsection{New Feature Synthesis via Linear Interpolation.}
\begin{figure}[t]
    \begin{minipage}{1.0\linewidth}
    \centering
    \includegraphics[width=8cm]{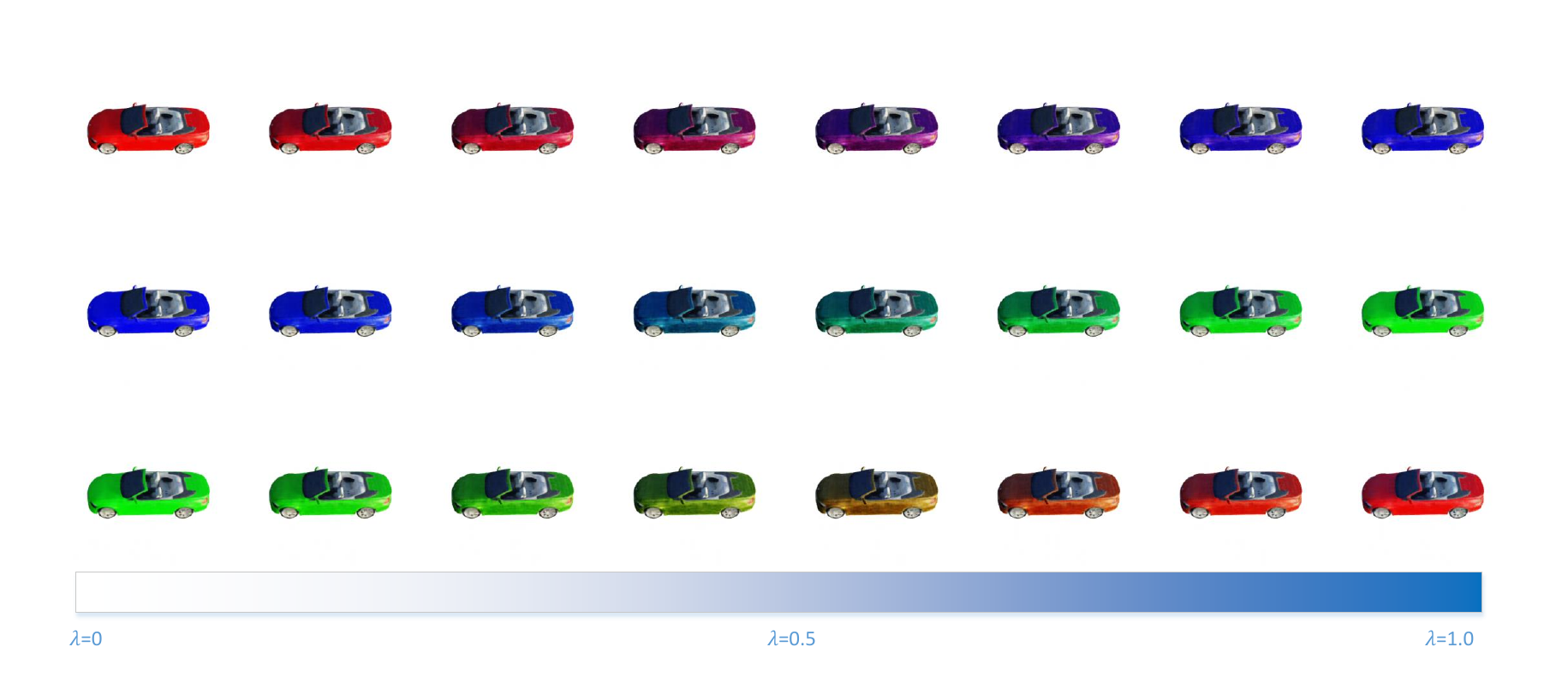}
    \caption{The color of the car is synthesized via color linear interpolation. $\lambda$ is a linear coefficient that varies from 0 to 1. }
    \label{fig:16}
    \end{minipage}
    \begin{minipage}{1.0\linewidth}
    \centering
    \includegraphics[width=8cm]{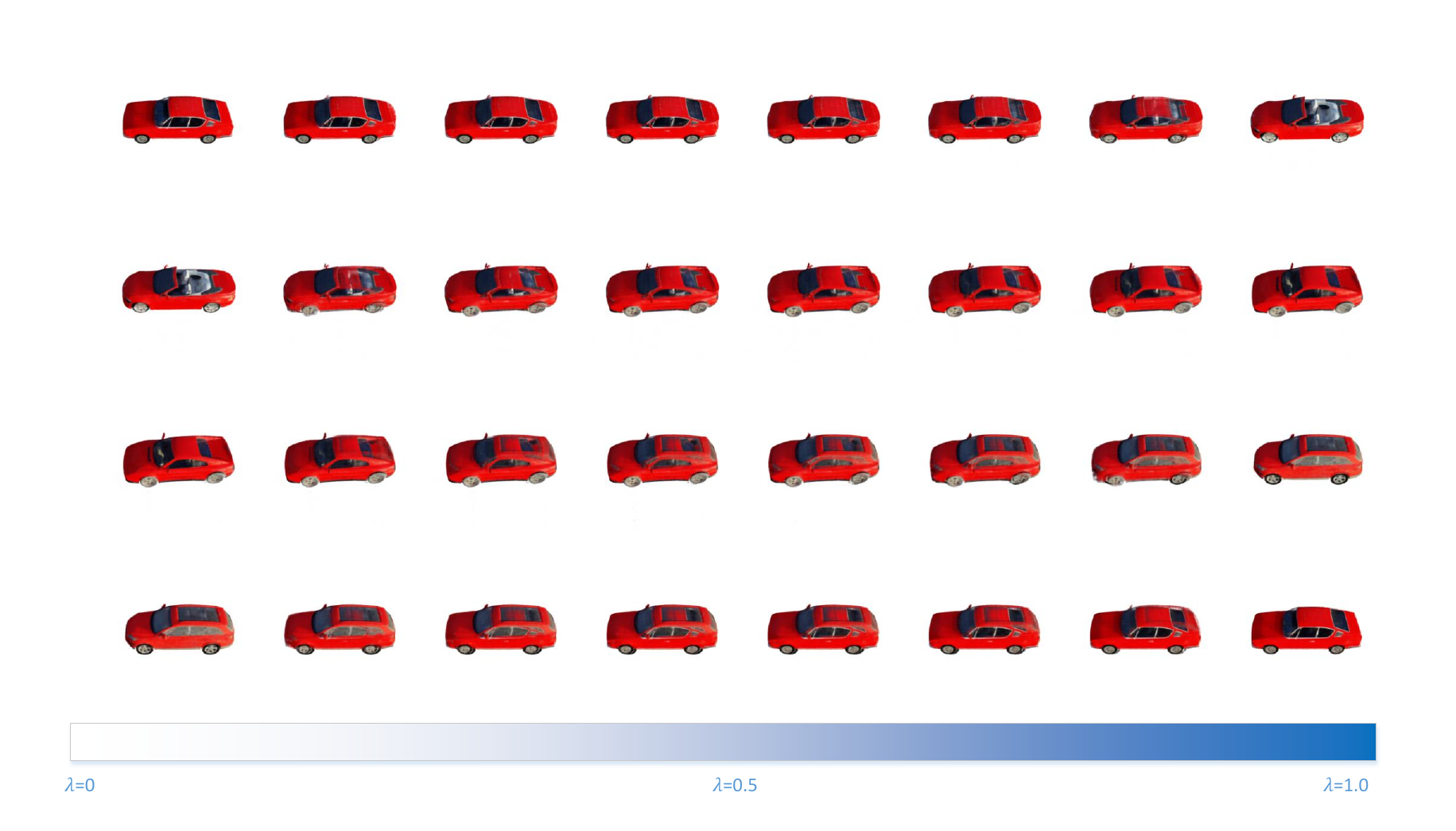}
    \caption{The shape of the car is altered via density linear interpolation. $\lambda$ is a linear coefficient ranging from 0 to 1.}
    \label{fig:17}
    \end{minipage}
\end{figure}

As shown in \textbf{Fig.}\ref{fig:16}, the new color of the car, which is unseen in the training set, is synthesized using the color interpolation: c=(1.0-$\lambda$)c[i]+$\lambda$c[j]. where $\lambda$ is a linear coefficient ranging from 0 to 1. In the same way, we can also simulate other features, such as texture, material, and environmental illumination. As shown in \textbf{Fig.}\ref{fig:17}, the shape of the car can also be altered step by step through density interpolation.
\subsection{Ablation Studies}
The development of the proposed model entailed adaptation of the GRAF prototype by altering the input and output components of the MLP and integrating an additional discriminator. Consequently, in our ablation studies, we compared the results of our model with those of Models I, II, and III, which eliminated the specific modifications for the input, output, and VGG discriminators. In Table \ref{tab:1}, we present a quantitative comparison of the FID and KID scores of Models I, II, and III with those of our model, indicating that the manipulation of the MLP output in the GRAF prototype plays a significant role in our model because training does not converge without it. Using density and color arrays, we effectively deployed the output to multiple slots corresponding to classes and styles. Then these outputs were used to render the images independently. We also observed that the image quality degraded without embedding labels for the input and the VGG discriminator. The two strategies not only increased the quality of the generated images, but also shortened the training time.
\begin{table*}[htbp]
\caption{\bf The results of FID/KID scores in the ablation studies.} 
\begin{tabular}{c c c c c c}
\toprule\\
\textbf{Model/Class}&Classic&Sporty&Roadster&Wagon&\textbf{Mean}\\
FID/KID&\\
\hline
\textbf{Model I}&$68.83$ & $71.07$ &$80.18$& $74.92$& $\textbf{73.75}$\\
(w/o input)&$0.072$ & $0.083$ &$0.078$& $0.075$& $\textbf{0.077}$
\\\midrule
\textbf{Model II}&$-$ & $-$ &$-$& $-$& no \\
(w/o output)&$-$ & $-$ &$-$& $-$& converge
\\\midrule
\textbf{Model III}&$62.13$ & $55.25$ &$58.14$& $63.96$& $\textbf{56.87}$ \\
(w/o VGG)&$0.065$ & $0.049$ &$0.060$& $0.068$&  $\textbf{0.061}$
\\\midrule
\textbf{Ours}&$42.54$ & $48.06$ &$51.43$& $44.37$&$\textbf{46.60} \downarrow $\\
&$0.048$ & $0.052$ &$0.057$& $0.045$&$\textbf{0.050} \downarrow $ 
\\
\bottomrule
\end{tabular}
\label{tab:1}
\end{table*}
\subsection{Comparison to SOTA Methods}
\begin{figure*}[h]
    \centering
    \includegraphics[width=13cm]{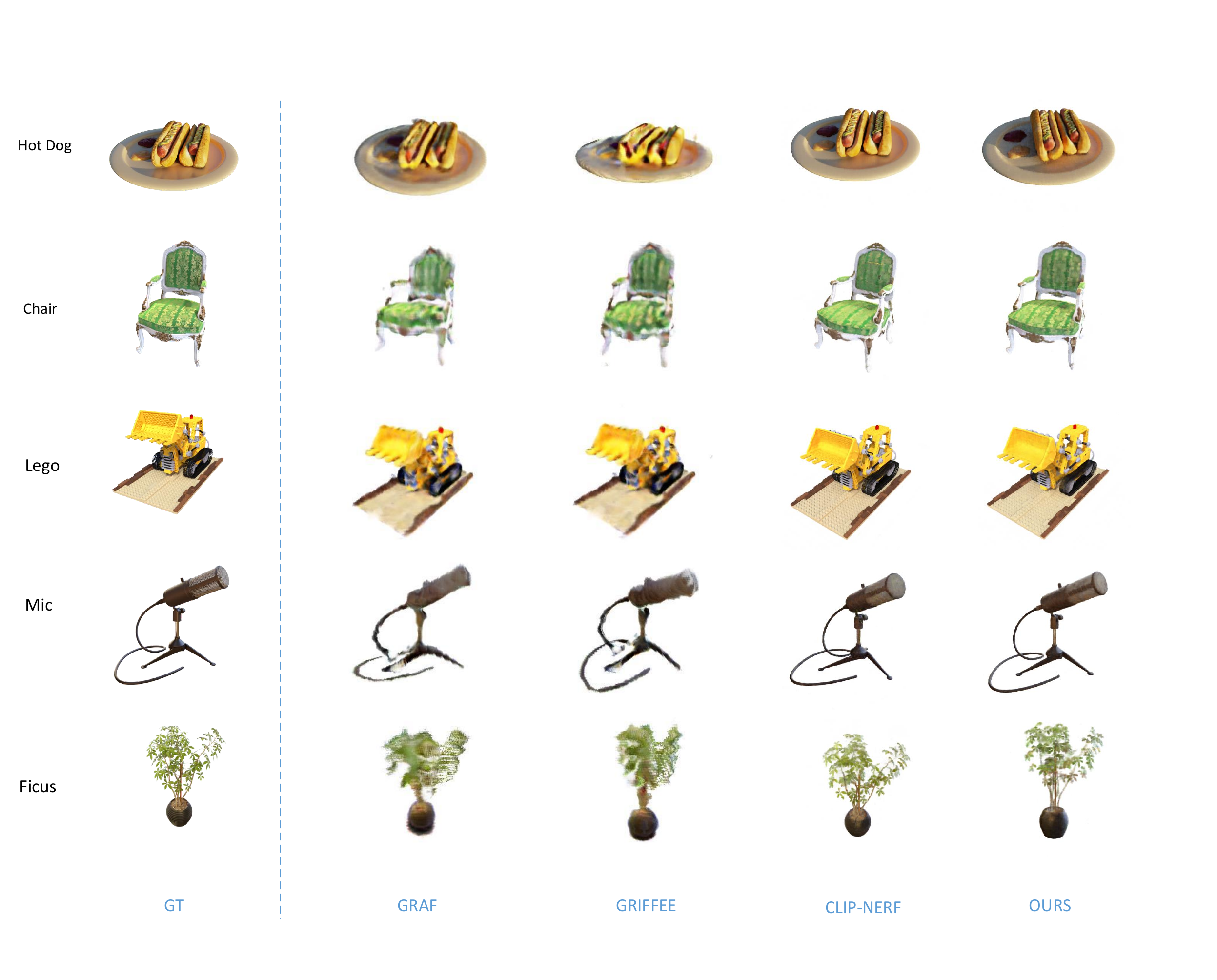}
    \caption{Some synthesized images by GRAF, GIRAFFE, CLIP-NERF and our model on the \textbf{Synthetic}(II) dataset.}
    \label{fig:18}
\end{figure*}

Generative radiance fields combine GAN and NERF techniques to synthesize 3D-aware images. In both qualitative (\textbf{Fig}. \ref{fig:18}) and quantitative (\textbf{Tab}.
\ref{tab:2}) comparisons with state-of-the-art generative methods, our approach yields results on par with the CLIP-NERF method and exceeds the GRAF and GIRAFFE methods in terms of PSNR and SSIM. During the experiment, we noticed that the generative models facilitate creating new views with 3D consistency using the unposed dataset, but our method can store multiple scenes in a single MLP without significantly sacrificing image quality and also allowing for manipulation of 3D-aware image creation based on the given labels and camera pose. Although GRAF and GIRAFFE are both generative models, GRAF is unable to represent multiple scenes within a single MLP. However, GIRAFFE employs MLPs to represent each object in a composite scene, leading to substantial memory consumption in multiobject scenes. As anticipated, our model falls short of CLIP-NERF with respect to PSNR and SSIM due to the use of a single MLP for implicit representation of multiple scenes simultaneously. Furthermore, to highlight the advantages of our model over CLIP-NERF, we performed a quantitative analysis in \textbf{Tab}. \ref{tab:3}, concentrating on storage requirements and computational costs. CLIP-NERF allows for the manipulation of the shape and color of objects according to text or image prompts, but it cannot learn multiple scene representations in a single model and requires separate training for each scene. As the number of scenes to be represented grows, both the model storage and the training time expand proportionally. In contrast, the demands of our model remain unchanged.

\begin{table*}[t]
\caption{\textbf{The quantitative assessment of our model with state-of-the-art methods in terms of PSNR and SSIM.}}
\centering
\begin{tabular}{lcccccccc}
\toprule
\textbf{Model}&\multicolumn{2}{c}{\textbf{\underline{GRAF\cite{2020GRAF}}}}&\multicolumn{2}{c}{\textbf{\underline{GIRAFFE\cite{2021GIRAFFE}}}}&\multicolumn{2}{c}{\textbf{\underline{CLIP-NERF\cite{2021CLIP}}}}&\multicolumn{2}{c}
{\textbf{\underline{OURS}}}\\
\textbf{/Class}&PSNR&SSIM&PSNR&SSIM&PSNR&SSIM&PSNR&SSIM \\
\midrule
Hot dog&22.17&0.980&22.67&0.981&34.38&0.998&32.21&0.997 \\
Chair&23.76&0.983&21.85&0.976&32.63&0.997&30.76&0.996 \\
Lego&24.53&0.986&23.16& 0.985&30.45&0.996&28.36& 0.995\\
Mic&21.48&0.976&22.64&0.980&31.38&0.996&29.57&0.995\\
Ficus&23.50&0.985&21.58&0.972&32.86&0.997&31.85&0.996\\
\midrule
Mean&\textbf{23.08}&\textbf{0.983}&\textbf{22.38}&\textbf{0.979}&\textbf{32.34}&\textbf{0.997}&\textbf{30.55} &\textbf{0.996} \\
\bottomrule
\end{tabular}
\label{tab:2}
\end{table*}

\begin{table*}[t]
\caption{\textbf{The comparison of our model with CLIP-NERF \cite{2021CLIP} in storage demands and computational expenses.}}
\centering
\begin{tabular}{lccccccccc}
\toprule
\textbf{Model}&\multicolumn{4}{c}{\textbf{\underline{CLIP-NERF}}} &&\multicolumn{4}{c}
{\textbf{\underline{OURS}}}\\
\textbf{/Num. of Scenes} (n) &n=1&n=2&n=3&n=4&&n=1&n=2&n=3&n=4\\
\midrule
\textbf{Model Storage} (MB) &13.6&27.2&40.8&54.4&&14.1&14.1&14.1&14.1 \\
\textbf{Training time} (hours)&32.6&65.8&98.5&131.2&&8.3&8.1&8.4&8.5 \\
\textbf{Inference time} (seconds)&7.75&7.58&7.61& 7.63&&7.63&7.78&7.68& 7.71\\
\bottomrule
\end{tabular}
\label{tab:3}
\end{table*}
\section{Conclusion}
The study aimed to achieve a sophisticated level of control in 3D-aware image synthesis. To achieve this goal, we improved the GRAF to allow for precise manipulation of 3D object creation in terms of pose, class, color style, and other attributes. By modifying the input and output of the MLP as well as the incorporation of an additional discriminator, we successfully entangled and disentangled the label codes into and out of the latent code, thereby enabling 3D-aware image generation from label prompts during the inference phase. Using our model, various scenes can be implicitly represented using a single MLP with shared weights, which significantly minimizes memory usage when handling multiple scenes. Additionally, it has been demonstrated that while the image quality produced by our model surpasses that of the NeRF-based generative models, it is marginally less impressive than CLIP-NERF, which is attributed to the shared weights within the MLP. Another limitation of the model is that the image quality diminishes as the quantity and complexity of the scenes increase. 
\bibliographystyle{ieee_fullname.bst}

\end{document}